\documentclass[nohyperref]{article}

\usepackage{microtype}
\usepackage{graphicx}
\usepackage{subfigure}
\usepackage{booktabs} %

\usepackage{hyperref}

\usepackage[accepted]{icml2023}

\usepackage{amsmath}
\usepackage{amssymb}
\usepackage{mathtools}
\usepackage{amsthm}
\usepackage{bbm}
\usepackage[capitalize,noabbrev]{cleveref}

\theoremstyle{plain}

\theoremstyle{definition}

\theoremstyle{remark}

\DeclareMathOperator*{\E}{\mathbb{E}}

\usepackage{enumitem}
\usepackage[textsize=tiny]{todonotes}
\usepackage{bm}

\def\eqref#1{equation~\ref{#1}}

\def\1{\bm{1}}

\DeclareMathAlphabet{\mathsfit}{\encodingdefault}{\sfdefault}{m}{sl}
\SetMathAlphabet{\mathsfit}{bold}{\encodingdefault}{\sfdefault}{bx}{n}

\icmltitlerunning{Mitigating Spurious Correlations in Multi-modal Models during Fine-tuning}

\begin{document}

\twocolumn[
\icmltitle{Mitigating Spurious Correlations in Multi-modal Models during Fine-tuning}

\icmlsetsymbol{equal}{*}

\begin{icmlauthorlist}
\icmlauthor{Yu Yang}{ucla}
\icmlauthor{Besmira Nushi}{msr}
\icmlauthor{Hamid Palangi}{msr}
\icmlauthor{Baharan Mirzasoleiman}{ucla}
\end{icmlauthorlist}

\icmlaffiliation{ucla}{Department of Computer Science, University of California, Los Angeles, USA}
\icmlaffiliation{msr}{Microsoft Research, Redmond, USA}

\icmlcorrespondingauthor{Yu Yang}{yuyang@cs.ucla.edu}
\icmlcorrespondingauthor{Besmira Nushi}{besmira.nushi@microsoft.com}

\icmlkeywords{Machine Learning, ICML}

\vskip 0.3in
]

\printAffiliationsAndNotice{}  %

\begin{abstract}
Spurious correlations that degrade model generalization or lead the model to be right for the wrong reasons are one of the main robustness concerns for real-world deployments. However, mitigating these correlations during pre-training for large-scale models can be costly and impractical, particularly for those without access to high-performance computing resources. This paper proposes a novel approach to address spurious correlations during fine-tuning for a given domain of interest. With a focus on multi-modal models (e.g., CLIP), the proposed method leverages different modalities in these models to detect and explicitly set apart spurious attributes from the affected class, achieved through a multi-modal contrastive loss function that expresses spurious relationships through language. Our experimental results and in-depth visualizations on CLIP show that such an intervention can effectively i) improve the model's accuracy when spurious attributes are not present, and ii) directs the model's activation maps towards the actual class rather than the spurious attribute when present. In particular, on the Waterbirds dataset, our algorithm achieved a worst-group accuracy 23\% higher than ERM on CLIP with a ResNet-50 backbone, and 32\% higher on CLIP with a ViT backbone, while maintaining the same average accuracy as ERM\footnote{Code can be found at \href{https://github.com/BigML-CS-UCLA/CLIP-spurious-finetune}{https://github.com/bigml-cs-ucla/clip-spurious-finetune}}.
\end{abstract}

\section{Introduction}
\label{sec:intro}
Vision-Language models (e.g., CLIP, DALL-E, Stable Diffusion, Imagen) are becoming pervasive in real-world deployments and have transformed the way large-scale model architectures are trained and used in different applications. Their multi-modal nature has not only enabled a large variety of tasks (e.g. text-to-image generation, visual question answering, image captioning) but is also facilitating better learning techniques that take advantage of data in several modalities to jointly learn embeddings that can then be reused in downstream tasks~\cite{radford2021learning,kamath2021mdetr,li2022grounded, vlp_aaai}. 

\begin{figure}[t]
  \centering
  \includegraphics[width=.8\linewidth]{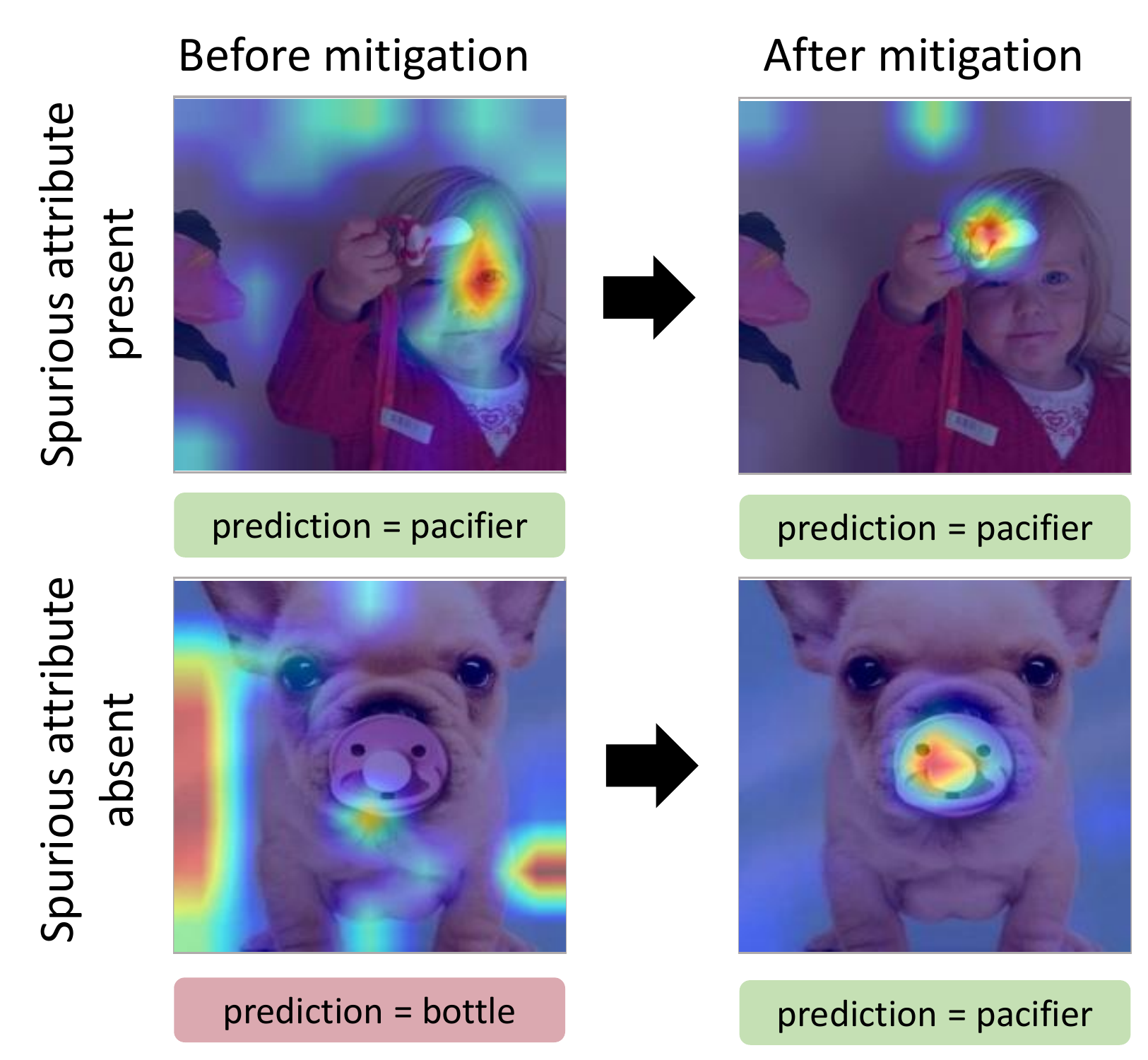}
  \caption{The \emph{baby pacifier} class in ImageNet is spuriously correlated with the presence of \emph{babies}, which leads the pre-trained model to be less accurate for cases when babies are absent in the image (bottom row) and also be right for the wrong reasons when babies are present (top row). Our approach mitigates both concerns by conveniently expressing and decorrelating the spurious relationships in the loss function via language.}\vspace{-2mm}
  \label{fig1}
\end{figure}

While the multi-modal alignment increases the expectations about model reliability due to better grounding and larger availability of data in general, these models are still not immune to fundamental learning problems such as dealing with spurious correlations~\cite{bommasani2021opportunities,moayeri2022comprehensive,petryk2022guiding,agarwal2021evaluating}. Therefore, when such models are used as a backbone to solve application-oriented tasks on a given domain, existing spurious correlations specific to that domain or the fine-tuning data that comes with it, may resurface in ways that are harmful to end users. At the same time, retraining large models from scratch to address such issues has become a less realistic avenue for two main reasons. First, stakeholders who  need to adapt a model to a particular domain may not necessarily have access to large-scale computation. Second, the types of spurious correlations of interest are often domain-specific and not all of them can be anticipated ahead of time during pre-training of a general model. Furthermore, while previous work has studied spurious correlations in single-modal models trained with supervised learning, we note that spurious correlations learned in a joint multimodal embedding space with contrastive language image pretraining may not be the same due to differences in inputs and training objectives. For instance, we found that certain spurious correlations commonly studied in supervised learning of vision models, such as the correlation between gender and hair colors in the CelebA dataset \cite{liu2015deep}, were not learned by multimodal models with contrastive language image pretraining. This suggests that spurious correlations in multimodal models may exhibit unique characteristics that require further investigation. 

Building on the challenges of spurious correlations in vision-language models and the need for efficient mitigation methods, we introduce a contrastive learning approach that leverages the multi-modality of CLIP as a vision-language model to detect and mitigate spurious correlations through language in fine-tuning time. In the detection stage, our method extracts linguistic attributes from the image and tests whether their presence or absence affects model performance. If the accuracy of the model drops when a specific attribute is not present, it indicates that the attribute is either an \emph{overemphasized but necessary attribute} (e.g., misclassifying taxi cabs that are not yellow) or a \emph{spurious correlation} (e.g., misclassifying boats when there is no water in the background)~\cite{singlaCVPR2021}. Assuming that a practitioner or domain expert in the loop can determine whether the attribute is healthy or spurious, in the next stage, our method mitigates the identified spurious correlation by extending the current contrastive language-vision learning techniques with a set of additional loss functions that explicitly i) decorrelate spurious attributes from the class names in language, and ii) push away both the vision representations across classes and language representations of templates substituted with different class labels. It is worth noting that our approach \emph{only fine-tunes the projections to the joint embedding space}. Since the projection layers contain much fewer parameters than the full models, our method requires significantly less computational resources compared to extensive retraining from scratch without losing features learned in pretraining.

In contrast to previous work which requires human annotations about spurious or group attributes~\cite{sagawa2019distributionally}, our approach uses automatically detected language-based descriptions of spurious attributes that can then directly be expressed and used in optimization to set them apart from affected classes. While domain experts are still required in this method to judge whether a detected co-occurence is a spurious attribute or not, this still minimizes labeling human supervision per example. Fine-tuning experiments with two datasets, Waterbirds and Imagenet, show that the proposed approach offers a better trade off between the average accuracy and worst-group accuracy (i.e., examples when the spurious attribute is not present) and can better align model explanation maps to the class of interest. 

It is worth noting that our work differs from existing studies that focus on spurious correlations learned by vision models \cite{sagawa2019distributionally,nam2020learning,creager2021environment,liu2021just,nam2022spread,izmailov2022on}. Instead, we investigate spurious correlations learned by multimodal models during pre-training with the contrastive language-image loss. Although larger models may be less accurate than specialized models on certain tasks, practitioners may still choose to use a pretrained model for reasons such as maintenance and data availability. In addition, having enough labeled data to train a specialized vision model may not always be possible. In such cases, the larger pretrained model trained on noisy image-caption pairs may have already encoded useful information about the concept, and our method is useful for scenarios where one needs to maintain this generality while mitigating found issues for a specific domain.

Moreover, the multimodal nature of these models opens up new opportunities for detecting and mitigating failures without the need for additional annotation data, such as attributes or bounding boxes, to guide the model's attention. By leveraging the information encoded in the joint embedding space, our approach improves the model's attention in GradCAM explanations and quantitatively in AIoU scores, a new metric we proposed for evaluating the model's attention. This finding is particularly noteworthy as the need for metadata annotations and grounding has been a significant barrier for several applications, especially during cold starts.

In summary, our contributions are:
\begin{itemize}[leftmargin=*,itemsep=0.0ex]
    \item A language-based approach that detects spurious correlations with practitioner supervision but no spurious attribute labeling.
    \item A loss function that extends current contrastive vision-language learning for mitigating spurious correlations in vision through language.
    \item A set of experiments that showcase how to use the proposed detection and mitigation approach in practice for the CLIP model as well as its effectiveness in datasets with known  and unknown spurious correlations.
\end{itemize}

\section{Related Work}
\label{sec:related}

\paragraph{Explaining and Debugging Trained Models.}
Several algorithms have been proposed to semantically explain and analyze trained neural networks, including distilling the decision modes into decision trees \cite{zhang2019interpreting,singlaCVPR2021}, training classifiers in the latent space \cite{jain2022distilling,yang2022explaining}, and embedding inputs with joint vision-language representations to find the error slices with a mixture model \cite{eyuboglu2022domino}. These methods usually accompany the semantic explanations with feature attention maps, e.g., GradCam \cite{selvaraju2017grad}. The authors of \cite{pmlr-v119-shankar20c} conducted a comprehensive study on ImageNet \cite{ILSVRC15} by manually relabeling it and uncovered multiple instances of label noise and disagreement in the dataset.
In this paper, we are only interested in discovering and mitigating \textit{spurious correlations}, which are introduced next.

\textbf{Enhancing Robustness to Spurious Correlations.}
We study spurious correlations in the context of deep learning, as they have been formally discussed in \cite{sagawa2019distributionally}. Given a classification dataset $\mathcal{D}$ with labels $\mathcal{Y}$, if there exist spurious attributes $\mathcal{A}$ that are highly correlated with $\mathcal{Y}$, a deep neural network trained on this dataset is likely to learn $\mathcal{A}$ as features to distinguish $\mathcal{Y}$, even if the attribute is not conceptually part of the class concept. 
For example, in \cref{fig1}, a pretrained CLIP-RN50 model \cite{radford2021learning} learned to use \textit{baby} to identify \textit{baby pacifier} because they often appear together in ImageNet \cite{ILSVRC15}, instead of actually learning the baby pacifier itself.

To prevent deep learning models from learning such spurious correlations from biased data, recent work proposed training strategies robust to spurious attributes for either vision or language models \cite{sagawa2019distributionally,nam2020learning,creager2021environment,liu2021just,nam2022spread,izmailov2022on}. The spurious label of each training example (e.g., whether this example contains the spurious feature) is either provided \cite{sagawa2019distributionally,izmailov2022on} or inferred by training a reference model \cite{nam2020learning,creager2021environment,liu2021just,nam2022spread} until it learns the spurious correlations. Other approaches indirectly estimate and use the causal effect of hidden non-labeled spurious attributes in pre-training~\cite{mao2022causal}.

However, these studies all focus on training \textit{unimodal} models with datasets that contain known spurious features, and spurious correlations learned by pretrained \textit{multimodal} models have not been extensively studied. To the best of our knowledge, we are the first to propose a \textit{fine-tuning} approach for mitigating spurious correlations in multimodal models. While \cite{zhang2022contrastive} also studied CLIP's robustness to group shifts including spurious correlations, their method is designed for transfer learning rather than fine-tuning the learned embedding space.

\textbf{Correcting Vision Models using Language.}
There is a line of recent work aiming to fix vision classifiers with language inputs. \citet{petryk2022guiding} uses attention maps from a pre-trained CLIP to supervise a CNN classifier's spatial attention. \citet{zhang2023diagnosing} probes a vision classifier trained on the joint vision-language embedding space of CLIP using language embeddings of attributes, identifies the attributes causing most failures, and generates a large set of natural language inputs with the influential attributes to rectify the model. However, this line of work aims to guide CNN classifiers rather than fixing CLIP models and does not prevent spurious feature usage.
\section{Spurious-aware Contrastive Language Image Fine-tuning}
\label{sec:method}

\paragraph{Background.}  Contrastive Language-Image Pretraining (CLIP) learns from millions of image caption pairs, by maximizing the agreement between representations of every image and the representations of its corresponding caption. 
Specifically, the CLIP architecture consists of (i) an image encoder network, %
(ii) a text encoder network, %
and (iii) a contrastive objective that pulls the embeddings of
every image and its corresponding caption together while pushing apart embeddings of the image from other captions in the same minibatch.
Formally, for a minibatch of $N$ image-captions pairs $\{I_j,T_j\}_{j=1}^N$, and their encoded embeddings $\{I_j^e,T_j^e\}_{j=1}^N$, the CLIP loss is defined as follows:
\begin{align}
    \mathcal{L}_{\text{CLIP}}
    =&-\frac{1}{2}\E_{(I_i,T_i)} \log \left [ \frac{e^{\left<I_j^e,T_j^e\right>/\tau}}{\sum_{k=1}^N e^{\left<I_j^e,T_k^e\right>/\tau} } \right] \label{eq:cliploss}\\
    &-\frac{1}{2}\E_{(I_i,T_i)} \log \left [ \frac{e^{\left<I_k^e,T_k^e\right>/\tau}}{\sum_{j=1}^N e^{\left<I_j^e,T_k^e\right>/\tau} } \right],\nonumber   
\end{align}
where $\left<.,.\right>$ represents the inner product, and $\tau$ is a trainable temperature parameter.
For finetuning CLIP on a dataset of images and their labels, such as Waterbirds, the labels are replaced in the engineered prompt templates, such as
``A photo of a \{label\}", 
``A photo of a \{label\}, a type of bird.”, etc. Then, the loss is minimized on the images paired with templates built with image labels. We use all 80 templates described in \cite{radford2021learning}.

For a given spurious attribute (e.g. water or land background in the Waterbirds dataset), we will use the following losses to eliminate the spurious correlation during fine-tuning. Please note that the contrastive losses below use the class information to pull together representations of examples from the same class label, and push away representations of examples from different class labels. The spurious losses use the spurious attribute detected in the spurious correlation detection stage (Section~\ref{sec:detection}) to pull together representations of examples with the same spurious attribute (e.g. attribute present) and push away representations of examples with a different spurious attribute (e.g. attribute absent). 

Here, we will use the following construct as a basis for the definition of all loss terms: a \emph{cross-group representation similarity} term that pulls together representations from the same group and pushes away representations of different groups. The representations can be either in the vision or language space. We reuse this construct to extend CLIP contrastive learning to improve classification and also mitigate spurious correlations. Let $G_1=\{(I_p,T_p)\}_{p=1}^P$ be the set of examples in one group of examples in the minibatch, and $G_2=\{(I_q,T_q)\}_{q=1}^Q$ the set of examples in another group of the same minibatch, as defined by the relationship of a given example in the minibatch $(I_i, T_i)$ to these groups. Depending on the loss term, the relationship between examples can be either due to examples belonging to the same class or having the same spurious attribute value. Then, the cross-group representation similarity defined across two modalities of representation embeddings $A$ and $B$ is:
\begin{align}
    \textsc{CS}\!=\!-\hspace{-6mm}\E_{\substack{(I_i,T_i),\\ (I_p,T_p) \in G_1\\ (I_q,T_q) \in G_2}} 
    \!\!\!\left[\log \frac{ e^{\left<A_{i}^e,B_{p}^e\right>/\tau}}{\sum_{p=1}^P e^{\left<A_i^e,B_p^e\right>/\tau}+\sum_{q=1}^Q e^{\left<A_i^e,B_q^e\right>/\tau}} \right]\!\! \nonumber
\end{align}

\paragraph{Contrastive Image Loss}
The first term is a contrastive image loss which pulls together image representations of a class, and pushes away image representations of different classes in the vision model.
Let $G_l=\{(I_p,T_p)\}_{p=1}^P$ be the set of examples in the minibatch with the \textbf{same label} as example $(I_i,T_i)$, i.e., $T_i\!=\!T_p$, and $\hat{G}_{l}=\{(I_q,T_q)\}_{q=1}^Q$ be the set of examples with a \textbf{different label}. Then the contrastive image loss within the vision representation embeddings I is defined as:
\begin{align}
    \mathcal{L}_{vc}\!=\!\hspace{-0mm} \textsc{CS}(G_l, \hat{G}_{l}, I, I)\!\! 
\end{align}

\paragraph{Contrastive Language Loss}
The second term is a contrastive language loss which pulls together language representations of templates of a class in the language model, and pushes away language representations of different classes.
Let $G_l=\{(I_p,T_p)\}_{p=1}^P$ be the set of examples in the minibatch with the \textbf{same label} as example $(I_i,T_i)$, i.e., $T_i\!=\!T_p$, but with different templates. Let $\hat{G}_{l}=\{(I_q,T_q)\}_{q=1}^Q$ be the set of examples in the minibatch with a \textbf{different label}. Then the contrastive language loss  within the language representation embeddings $T$ is defined as:
\begin{align}
    \mathcal{L}_{lc}\!=\!\hspace{-0mm} \textsc{CS}(G_l, \hat{G}_{l}, T, T)\!\! 
\end{align}

\paragraph{Spurious Image Loss}
The third term is a spurious contrastive image loss which pulls together image representations of each group of examples in a class, and pushes away image representations of different groups of examples. For example, it pulls together images of waterbirds with water background, and pulls them away from images of waterbirds with land background and from landbird images with water or land background.

Assume $G_s=\{(I_p,T_p)\}_{p=1}^P$ is the set of images in the minibatch with the \textbf{same spurious attribute} as example $(I_i,T_i)$, and $\{\hat{G}_s=(I_q,T_q)\}_{q=1}^Q$ is the set of examples with a \textbf{different spurious attribute} than example $(I_i,T_i)$.  A different spurious attribute here could also mean that the spurious attribute is absent. Then, the spurious image loss  within the vision representation embeddings I is defined as:
\begin{align}
    \mathcal{L}_{vs}\!=\!\hspace{-0mm} \textsc{CS}(G_s, \hat{G}_{s}, I, I)\!\! 
\end{align}

\paragraph{Spurious Language Loss}
The last term is a spurious contrastive language loss which pulls together language representations of each group of examples in a class, and pushes away language representations of different groups of examples.
Assume $G_s=\{(I_p,T_p)\}_{p=1}^P$ is the set of examples in the minibatch with the \textbf{same spurious attribute} with example $(I_i,T_i)$, and $\hat{G}_s=\{(I_q,T_q)\}_{q=1}^Q$ is the set of examples with a \textbf{different spurious attribute} in the minibatch. Note that, a different spurious attribute here could also mean that a spurious attribute is absent. Then, the spurious language loss  within the language representation embeddings T is defined as:
\begin{align}
    \mathcal{L}_{ls}\!=\!\hspace{-0mm} \textsc{CS}(G_s, \hat{G}_{s}, T, T)\!\! 
\end{align}

The final loss is the sum of all the terms above:
\begin{equation}
\mathcal{L}=\mathcal{L}_{\text{CLIP}}+\mathcal{L}_{vc}+\mathcal{L}_{lc}+\mathcal{L}_{vs}+\mathcal{L}_{ls}.
\end{equation}

In practice, either $\mathcal{L}_{vs}$ or $\mathcal{L}_{ls}$ can be combined with $\mathcal{L}_{lc}$ to effectively eliminate the spurious correlation. If spurious attribute annotation labels are available, one can use $\mathcal{L}_{vs}$. If spurious attribute annotation labels are not available $\mathcal{L}_{ls}$ can provide a good separation between groups in different classes. In all experiments reported hereafter we show results for both, and the ablation study in Section~\ref{sec:loss-ablation} details the tradeoffs between these and other choices.

From an implementation perspective, all language-related losses could be implemented across examples or templates. Our implementation follows a template-based approach.

\begin{figure*}[ht]
\vskip 0.2in
\begin{center}
\centerline{\includegraphics[width=.8\textwidth]{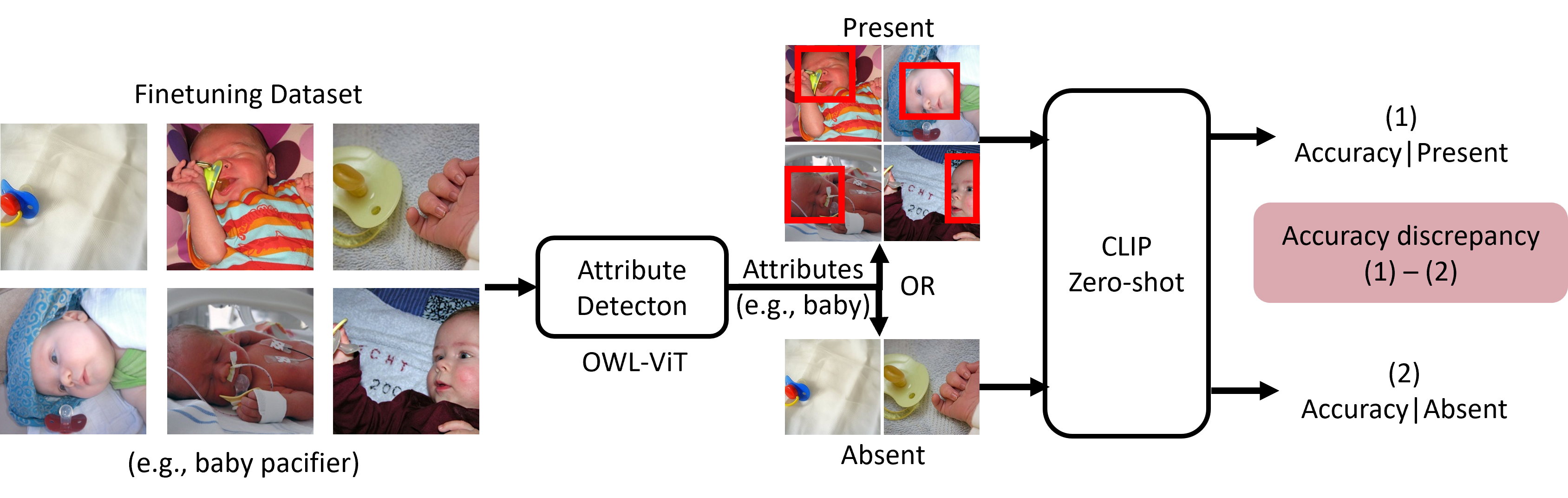}}
\caption{Spurious correlation detection based on attributes from an open-vocabulary detector and accuracy discrepancy scores of the model between examples when the spurious attribute is present or absent. }
\label{fig:detection-diagram}
\end{center}
\vskip -0.2in
\end{figure*}

\section{Spurious Correlation Detection}
\label{sec:detection}
This section introduces our pipeline for detecting and evaluating spurious correlations learned by a pretrained model. While we apply this pipeline to CLIP models in this work, it can be generalized to other pretrained models as well. The approach closely follows previously discussed techniques~\cite{singlaCVPR2021,nushi2018towards} but relies on automatically generated annotations for attributes.

\subsection{Methodology}

For any given fine-tuning dataset, we are interested in knowing whether CLIP (or any other pretrained models) has learned any spurious correlations for the classes in the dataset. According to the definition of spurious attributes introduced in \cref{sec:method}, models that have learned a certain spurious correlation usually show better performance (e.g., higher classification accuracy) on examples with that spurious attribute. For example, a model that majorly relies on the presence of an emergency vehicle to detect an accident, would have a lower accuracy in detecting accidents when there are no emergency vehicles around. 

We use the pipeline depicted in Figure~\ref{fig:detection-diagram} to (1) find such spurious attributes for a class of interest if the spurious attributes are unknown, and (2) measure how much each spurious attribute negatively affects the model. 

\paragraph{Spurious Detection.} For the case where the spurious attribute is unknown, we first use an open-vocabulary detector, OWL-ViT \cite{minderer2022simple}, to detect potential spurious attributes for examples in the fine-tuning data. We use the synsets of object names in Visual Genome \cite{krishnavisualgenome} as our list of attributes to detect after removing objects that are classes of the fine-tuning data. 

\paragraph{Spurious Evaluation.} We define $\delta(\mathcal{D}, s)$ as the model accuracy discrepancy between examples in dataset $\mathcal{D}$ with the attribute $s$ and those without it.  
\vspace{-1mm}
\begin{align}
\delta(\mathcal{D},s) = acc(\mathcal{D} | s=1) - acc(\mathcal{D} | s=0).
\end{align}
Attributes detected in the fine-tuning dataset can then be ranked by their accuracy discrepancy scores. The higher the discrepancy, the more this attribute could harm the generalization performance of the pretrained model. Since model failure modes and in particular spurious correlations are often specific to the class~\cite{nushi2018towards}, for the ImageNet studies we compute and sort the discrepancy scores per class. While the drop in accuracy with the absence of the spurious attributes are good indicators of spurious correlations, such drops may also happen for healthy attributes that are part of the class definition (e.g., the yellow color for taxi cabs albeit not all taxis are yellow).

Thus, for practical usages of our approach, we imagine this step to involve some miminal human investigation from domain experts or ML practitioners to judge whether the attribute is healthy or a potential spurious correlation. Humans can make this call based on their domain knowledge or one of the vision interpretability techniques (e.g, Grad-CAM, Integrated Gradients etc.). Nevertheless, this kind of supervision is considerably more lightweight than annotating attributes or manually inspecting individual examples. \cref{tab:imagenet_spurious_correlations} shows several examples of previously unknown spurious correlations we found for CLIP in ImageNet.

\section{Experiments}
\subsection{Backbones} CLIP uses two main groups of visual backbones, ResNets (RN) and Visual Transformers (ViT), and reported model performance separately for models with these two types of backbones in \cite{radford2021learning}. In particular, ResNet-50 (RN50) and ViT-L/14@336px \footnote{ViT-L/14@336px refers to ViT-L/14 model fine-tuned on 336-by-336 pixel input images.} are used as the prototypes of these two groups of models. Therefore, we follow \cite{radford2021learning} and study CLIP models with RN50 and ViT-L/14@336px visual backbones in our experiments.

For all experiments, we freeze both the language and vision encoders and only fine-tune the projection layers. Keeping both encoders intact is not only more lightweight but also resulted in better overall and worst-group accuracy for all studied datasets in our preliminary experiments. 
\begin{table}[t]
\caption{Statistics of the Waterbirds training data.}
\label{tab:waterbirds}
\vskip 0.0in
\begin{center}
\begin{small}
\begin{sc}
\begin{tabular}{lcc}
\toprule
 & Land & Water \\
\midrule
Landbirds  & 3498 & 184 \\
Waterbirds & \underline{56} & 1057 \\
\bottomrule
\end{tabular}
\end{sc}
\end{small}
\end{center}
\vskip -0.1in
\end{table}

\subsection{Datasets}

\paragraph{Waterbirds.} Waterbirds \cite{sagawa2019distributionally} is the most commonly used benchmark dataset for studying spurious correlations. It combines birds segmented from the CUB dataset \cite{WahCUB_200_2011} and the background in  dataset \cite{zhou2017places} in an imbalanced way such that the background can be used as a spurious attribute for bird classification. 
 \cref{tab:waterbirds} shows the sample size of each class-background combination in the Waterbirds training set. As landbirds appear more with land background and waterbirds are more often on water background in the training set, models fine-tuned on this dataset often learn to rely on the background instead of the birds.

\textbf{ImageNet-1K.}
\citet{singlaCVPR2021} found that some features are spuriously correlated with some categories in ImageNet-1K \cite{ILSVRC15}. For example, 55\% of training examples in the ``Rhodesian ridgeback" class can be correctly classified by a robust ResNet-50 model but the accuracy drops significantly to 24\% when the dogs are not wearing a collar. We use the spurious detection pipeline shown in \cref{fig:detection-diagram} to find top-5 attributes with the highest accuracy discrepancy on CLIP for each class and attribute, and then rank the top attributes from all classes. Based on our inspection of the top attributes, we find a number of previously unknown spurious attributes learned by CLIP-RN50 with ImageNet as shown in Table~\ref{tab:imagenet_spurious_correlations}. Out of this list, in the mitigation experiments we choose to mitigate the first major spurious correlation that has a high accuracy discrepancy: \emph{Baby pacifier} class where the spurious attribute is \emph{baby face}. CLIP accuracy drops by 69.1\% for classifying baby pacifiers when there is no baby in the image. Note that since the validation set for ImagenNet contains only 50 images per class, we run the spurious correlation detection and evaluation stages on the training data instead, while mitigation results are presented for the test data. 
Figures \ref{fig:imagenet-spurious} and \ref{fig:imagenet-rn50-mitigated} show further evidence of the pre-trained model focusing on the spurious attributes rather than the class itself.

Another dataset we considered for evaluation is CelebA~\cite{liu2015deep} for the task of hair color classification. Previous work~\cite{mao2022causal,sagawa2019distributionally} has shown that models trained on such data can have a lower accuracy for small groups defined by the gender attribute such as men with blond hair, since this group has a low representation in the training data. It turns out however that model accuracy does not degrade for this group using CLIP model, which is why we do not present results on CelebA in this paper.

\subsection{Metrics}
We use the following metrics to evaluate the \textbf{predictions} and \textbf{explanations} of each model. We argue that only by obtaining high performance in both aspects, an algorithm can be proven to address the spurious correlations and that the correct model predictions are ``\textit{right for the right reasons}''.  
\begin{enumerate}[leftmargin=*,itemsep=0.5ex]
    \item \textbf{Average Accuracy.} Classification accuracy averaged over classes on the test set. For the Waterbirds dataset, the test data is enriched and balanced to improve the accuracy of the evaluation, but this can lead to a discrepancy between the distribution of the test data and the training data. Following previous works, we report the adjusted average accuracy suggested by \cite{sagawa2019distributionally}, which weights the test accuracy of each group by their sizes in the training data. 
    \item \textbf{Worst-group Accuracy.} The lowest model accuracy across groups as defined by the spurious attribute and the class of interest. 
    \item \textbf{Adjusted Intersection-over-Union (AIoU).} Previous works have used binary attribute maps to compute an Intersection-over-Union (IoU) score with the ground-truth bounding box \cite{nguyen2021the}. While IoU is a standard metric for object localization, using the standard IoU to evaluate the quality of attribute maps can be less reliable because the score highly depends on the threshold used for binarizing the attribute maps. To circumvent threshold dependency, we adapt the formulation such that it instead uses a min operator (instead of the binary intersection) between a bounding box $B_y$ and an explanation map $M_y$, where $y$ is the ground truth class. Similarly, we use a max operator (instead of binary union) in the denominator between the bounding box and the map.
    \begin{align}
    \begin{split}
        \textsc{IoU}(M, B) = \frac{\sum_{j, k} \min(M_{jk}, B_{jk})}{\sum_{j, k} \max(M_{jk}, B_{jk})}, \\
        0 \le j \le h, 0 \le k \le w. 
    \end{split} 
    \label{eq:iou}
    \end{align}

    Equation~\ref{eq:iou} measures the alignment between an explanation map and the ground truth bounding box but it does not take into consideration that despite a good alignment with the bounding box for true class, the explanation maps of other classes may still span across the bounding box of the ground truth class. %
    Therefore, we use a definition of IoU that adjusts its denominator to include the class whose explanation map most intersects with the ground truth bounding box. 
    \begin{align}
    \begin{split}
        \hspace{-2mm}\textsc{AIoU} = 
        \frac{\textsc{IoU}(M_y, B_y)}{\textsc{IoU}(M_y, B_y)+ \underset{y'\in [C\setminus y]}{\max_{}} \textsc{IoU}(M_{y'}, B_y)}.
    \end{split} 
    \label{eq:aiou}
    \end{align}
In our experiments, we used GradCAM \cite{selvaraju2017grad} for the explanation maps. While GradCAM explanations may not be perfectly aligned with the model's attention, their usage has shown practical benefits for model debugging~\cite{YosinskiCNFL15,Simonyan2013DeepIC,mao2022causal}.
\end{enumerate}

\begin{table}[t]
\caption{Accuracy of different groups of Waterbirds on pre-trained ResNet- and Transformer-based CLIP models.}
\label{tab:waterbirds-pretrain}
\begin{center}
\begin{small}
\begin{sc}
\begin{tabular}{lcc}
\toprule
 (RN50) & Land & Water \\
\midrule
Landbirds & 93.44\%  & \underline{44.92\%}  \\
Waterbirds & 59.03\% & 91.59\% \\
\bottomrule
\end{tabular}
\vskip 0.15in
\begin{tabular}{lcc}
\toprule
 (ViT-L/14@336px) & Land & Water \\
\midrule
Landbirds  & 99.29\% & 90.20\% \\
Waterbirds & \underline{33.96\%} & 55.61\% \\
\bottomrule
\end{tabular}
\end{sc}
\end{small}
\end{center}
\vspace{-4mm}
\end{table}

\subsection{Baselines}
We compare our approach with pre-trained CLIP \cite{radford2021learning}, fine-tuned CLIP with the training dataset in hand using the original contrastive vision and language loss as described in Equation~\ref{eq:cliploss}, Empirical Risk Mimimization (ERM), and Group DRO~\cite{sagawa2019distributionally}. Group DRO is a distributionally robust optimization approach that minimizes worst-group loss and uses strong regularization. The methods requires attribute annotations to define groups being used during optimization. ERM instead is the standard empirical risk minimization technique for minimizing classification loss.

\textbf{Reproducibility.}
Both CLIP models (CLIP-RN50 and CLIP-ViT) and prompt templates we use in our experiments are officially released by OpenAI\footnote{https://github.com/openai/CLIP}\cite{radford2021learning}. The Waterbirds dataset is from the WILDS library \cite{wilds2021}. 
We used the SGD optimizer for all the experiemnts, and tuned the learning rates and weight decays for ERM, GroupDRO and CLIP-based loss (CLIP-finetuning and our method) separately. Our method uses learning rate 1e-5 with weight decay 1e-4. 
The code will be publicly available upon publication.

\begin{figure}[t]
\vskip 0.2in
\begin{center}
\centerline{\includegraphics[width=\columnwidth]{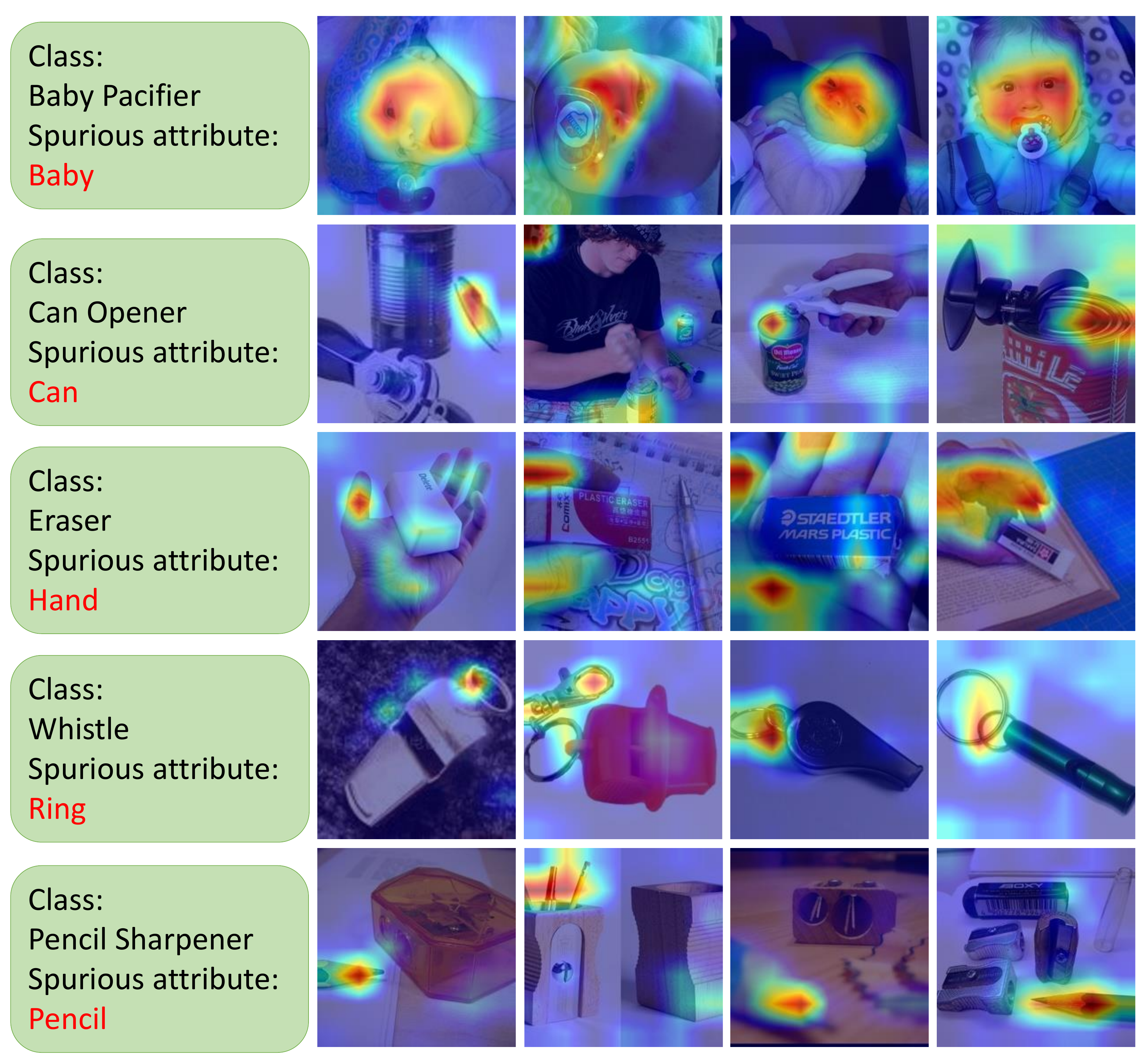}}
\vspace{-1mm}
\caption{GradCAM explanations for cases when Pre-trained CLIP RN50 relies on the spurious classification described in Table~\ref{tab:imagenet_spurious_correlations}.}
\label{fig:imagenet-spurious}
\end{center}
\vspace{-4mm}
\end{figure}

\begin{table}[t]
  \centering
    \caption{Spurious correlations found for CLIP RN50 on ImageNet.}
  \begin{tabular}{@{}lccc@{}}
    \toprule
    Class & Spurious  & Confused  & Acc. \\
     & Attribute & Class & Discrepancy\\
    \midrule
    baby pacifier & baby & water bottle & 62.1\% \\ 
    can opener & can & letter opener & 45.2\% \\ 
    eraser & hand & pencil case & 18.5\%  \\
    whistle & ring & padlock & 15.2\%\\ 
    pencil sharpener & pencil & pencil case & 8.37\% \\ 
         
    \bottomrule
  \end{tabular}
  \label{tab:imagenet_spurious_correlations}
\end{table}

\subsection{Spurious Correlation Detection Results}
\paragraph{Waterbirds.} \cref{tab:waterbirds-pretrain} shows model accuracy across the four groups as defined by class and spurious attribute definitions. The underlined groups show the worst-group accuracies for each model. For both models, there is a high accuracy discrepancy between groups from the same class. Figures~\ref{fig:waterbirds-rn50} and~\ref{fig:waterbirds-vit} show examples of explanations from Pre-trained CLIP where explanations do not overlap with birds.

\textbf{Imagenet.} Table~\ref{tab:imagenet_spurious_correlations} shows examples of prominent spurious correlations found for Pre-trained CLIP RN50. It is interesting to see how the found spurious attributes are concepts that are indeed highly related to the class but not necessarily part of the class definition. The natural co-occurence of these concepts leads the model to incorrectly rely rather on the attribute as shown in Figure~\ref{fig:imagenet-spurious}.

\subsection{Spurious Correlation Mitigation Results}
\begin{figure}[t]
\vskip 0.2in
\begin{center}
\centerline{\includegraphics[width=\columnwidth]{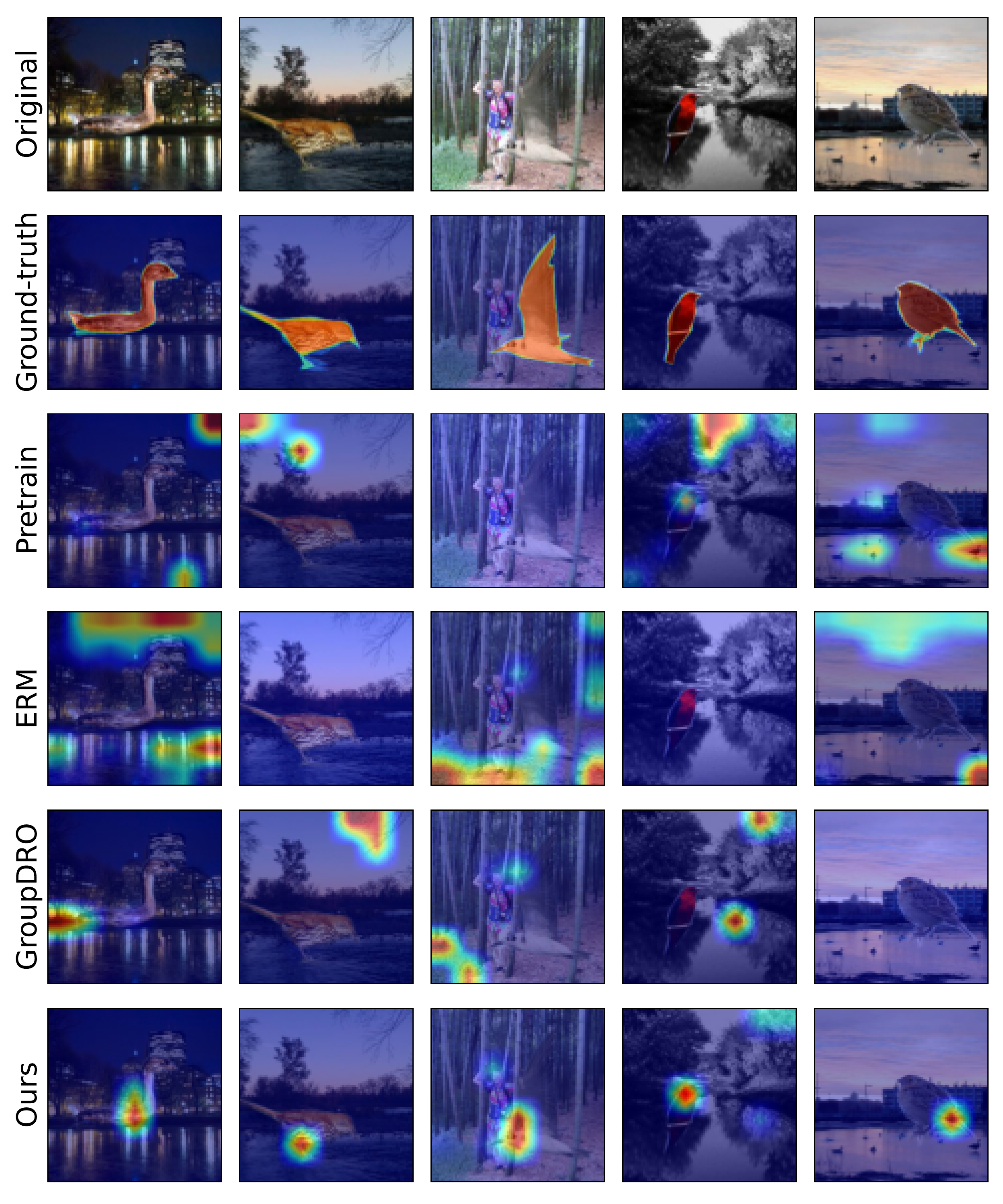}}
\vspace{-2mm}
\caption{GradCAM explanations for different approaches based on CLIP RN50 for the Waterbirds dataset.}
\label{fig:waterbirds-rn50}
\end{center}
\vspace{-5mm}
\end{figure}

\begin{figure}[t]
\vskip 0.2in
\begin{center}
\centerline{\includegraphics[width=\columnwidth]{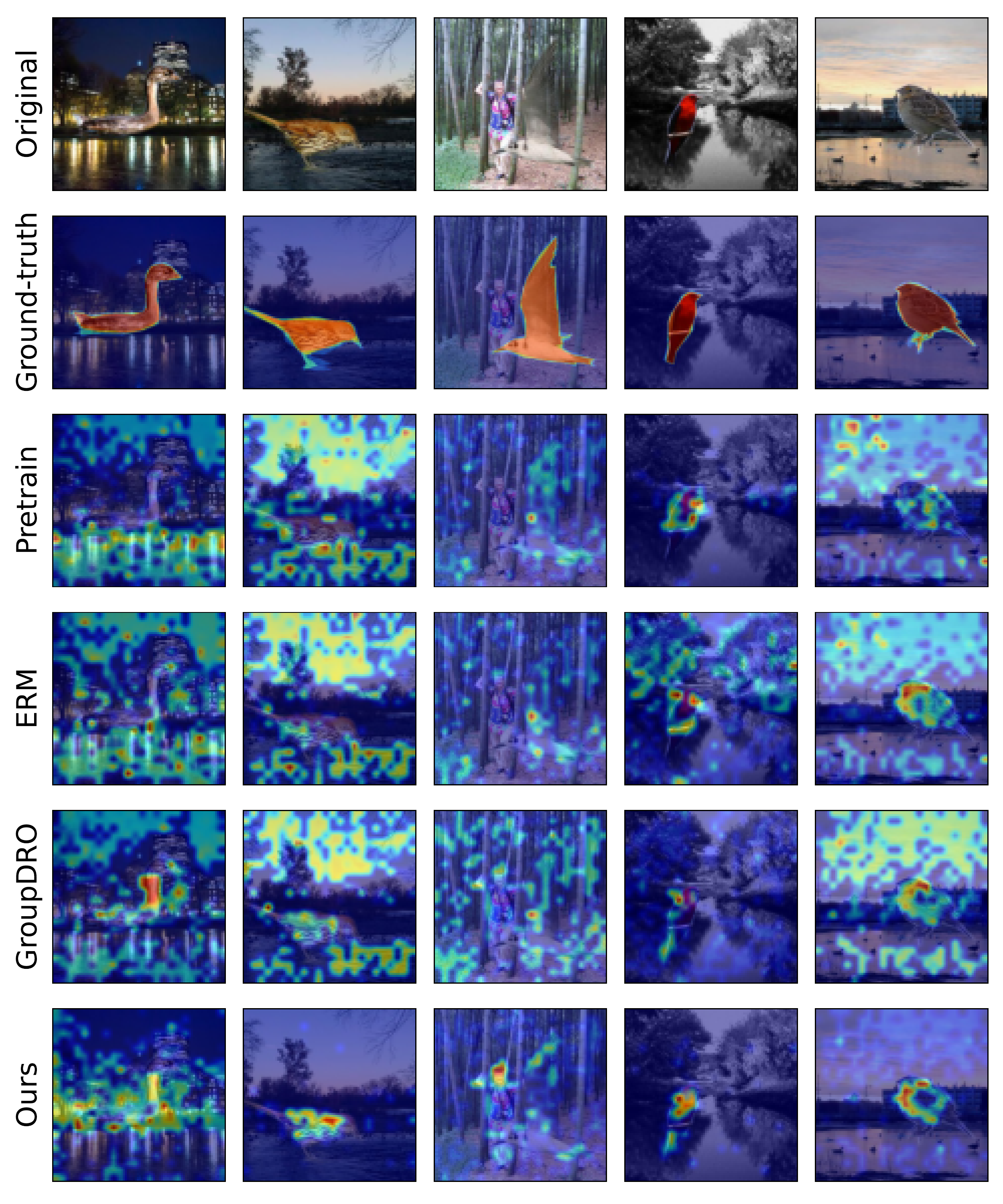}}
\vspace{-2mm}
\caption{GradCAM explanations for different approaches based on CLIP ViT-L/14@336px for the Waterbirds dataset.}
\label{fig:waterbirds-vit}
\end{center}
\vspace{-9mm}
\end{figure}

\begin{figure}[t]
\vskip 0.2in
\begin{center}
\vspace{-2mm}
\centerline{\includegraphics[width=\columnwidth]{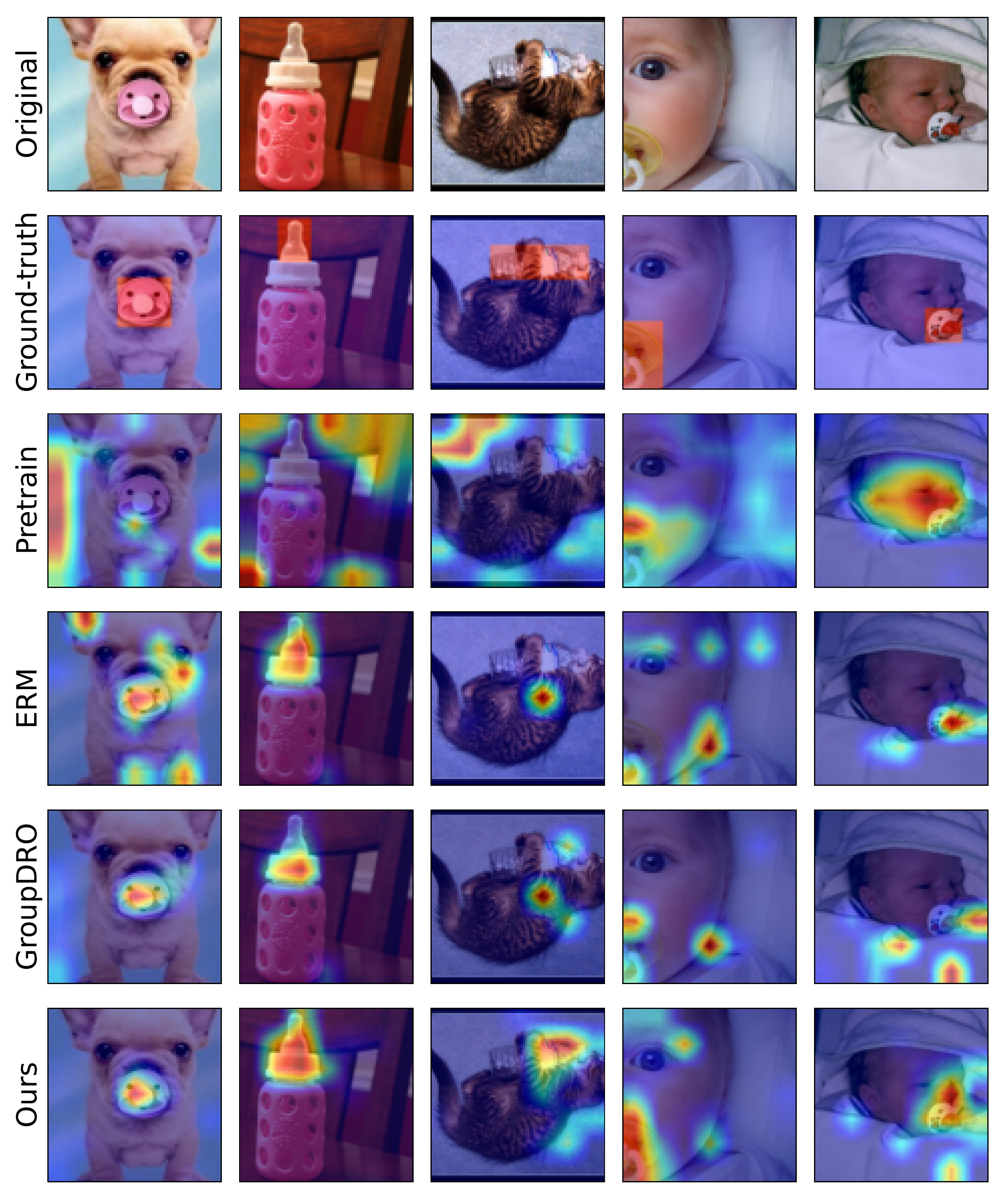}}
\vspace{-2mm}
\caption{GradCAM explanations for different approaches based on CLIP RN50 for the ImageNet dataset.}
\label{fig:imagenet-rn50-mitigated}
\end{center}
\vspace{-6mm}
\end{figure}

\begin{table*}[t]
  \centering
    \caption{Results of fine-tuning CLIP with Waterbirds. Average and worst-group performance is evaluated on the test set with models early stopped at the \textit{highest worst-group accuracy} on the validation set. Worst groups: \emph{Landbird on water} for RN50; \emph{Waterbird on land} for ViT.}
  \begin{tabular}{@{}lcccccccc@{}}
    \toprule
     Model & \multicolumn{4}{c}{ResNet-50}  & \multicolumn{4}{c}{ViT-L/14@336px} \\
     & \multicolumn{2}{c}{Accuracy} & \multicolumn{2}{c}{AIoU}  & \multicolumn{2}{c}{Accuracy} & \multicolumn{2}{c}{AIoU} \\
     & Avg. & Worst-group & Avg. & Worst-group  & Avg. & Worst-group & Avg. & Worst-group \\
    \midrule
    Pre-trained CLIP & \textbf{90.8\%} & 44.9\% & 0.507 & 0.479 & 88.5\% & 34.0\% & 0.579 & 0.551 \\
    Fine-tuned CLIP & 81.3\% & \textbf{77.1\%} & 0.510 & 0.128 & \textbf{97.2\%} & 89.7\% & 0.687 & 0.697\\
    \midrule
    ERM & \textbf{93.5\%} & 54.4\% & 0.514 & 0.139 & 96.8\%& 58.1\% & 0.636 & 0.680 \\
    Group DRO & 83.3\% & 73.7\% & 0.509 & 0.274 & 94.1\% & \textbf{90.8\%}  & 0.669 & 0.644 \\
    Ours($\mathcal{L}_{lc}$+$\mathcal{L}_{vc}$+$\mathcal{L}_{ls}$) & 84.7\% & \textbf{77.5\%} & \textbf{0.628} & \textbf{0.499} & \textbf{97.1\%} & 89.7\% & \textbf{0.698} & \textbf{0.711} \\
    Ours($\mathcal{L}_{lc}$+$\mathcal{L}_{vc}$+$\mathcal{L}_{vs}$) & 83.2\% & \textbf{77.5\%} & \textbf{0.654} & \textbf{0.587} & 96.9\% & \textbf{90.5\%} & \textbf{0.716} & \textbf{0.709} \\
    \bottomrule
  \end{tabular}
  \label{tab:waterbirdsresults}
  \vspace{-4mm}
\end{table*}
\paragraph{Waterbirds.} Table~\ref{tab:waterbirdsresults} summarizes our results on the Waterbirds dataset for both Resnet-50 and ViT-L/14@336px. Our method of mitigating spurious correlations through language has the best worst-group accuracy for ResNet-50 and second-best worst-group accuracy for ViT, maintaining a competitive average accuracy. What is of most interest from a mitigation perspective, is that the model ability to be right for the right reasons is indeed better for our method as indicated by the AIoU scores. These results are also qualitatively confirmed by visual explanation maps as shown in Figures~\ref{fig:waterbirds-rn50} and \ref{fig:waterbirds-vit}, demonstrating that (i) the spurious correlation is present on the first place (pre-trained CLIP), (ii) it persists in the explanation maps of GroupDRO despite this method being competitive in both worst-group and average accuracy, and (iii) it is visibly alleviated though our approach whose explanations align with the available ground truth segmentations for the dataset. When comparing the two different variants of our method using the spurious language loss and image loss, we observe that the spurious image loss leads to  better AIoU scores potentially because decorrelation is easier in the image representation, albeit for this method to work reliable attribute annotations are required. Using the spurious language loss is however still appealing with respect to both worst-group accuracy and AIoU. Note that implicitly, this method, and generally mitigating spurious correlations through language, relies on the capability of the model to map the spurious attribute from language to vision, which may not always be the case for the pre-trained vision-language models. The spurious attributes studied in this paper are based on the language concepts that are perhaps well-learned and mapped in a multi-modal way in CLIP (e.g., baby, water, land) but in other cases of less-frequent or domain-specific attributes, using the spurious image loss may be a more realistic avenue.

When comparing these findings between the two different model backbones we observe that AIoU scores for the ViT model are higher for all methods than their corresponding ResNet versions, indicating that the larger transformer-based model is perhaps more prone to improve upon using such mitigation techniques or even standard fine-tuning.

\begin{table}
  \centering
    \caption{Results of fine-tuning CLIP-RN50 with a subset of ImageNet classes, ``baby pacifier" and ``water bottle". Both average and worst-group performance are evaluated with models early stopped at the \textit{highest worst-group accuracy} on the validation set.}
  \begin{tabular}{@{}lcccc@{}}
    \toprule
     Class 680   & \multicolumn{2}{c}{Accuracy} & \multicolumn{2}{c}{AIoU}  \\
     Baby Pacifier & Avg. & Worst & Avg. & Worst \\
    \midrule
    Pre-trained & 73.7\% & 30.8\% & 0.651 & 0.380 \\
    Fine-tuned  &  94.1\% & 91.7\% & 0.650 & 0.571 \\
    \midrule
    ERM & \textbf{94.9\%} & \textbf{96.2\%} & 0.661 & 0.454  \\
    Group DRO & 89.6\% & 93.1\% & 0.661 & 0.568 \\
    Ours($\mathcal{L}_{lc}$+$\mathcal{L}_{vc}$+$\mathcal{L}_{ls}$) & \textbf{94.9\%} & \textbf{96.2\%} & \textbf{0.720} & \textbf{0.645}  \\
    \bottomrule
  \end{tabular}
  \label{tab:imagenetresults}
\end{table}

\textbf{ImageNet-1K.}
Here, we choose one of the most spurious correlations we found for the CLIP ResNet50: the \emph{baby pacifier} class where the spurious attribute is \emph{baby face}. The accuracy discrepancy between cases when there is a baby and no baby in the image is 69.2\% in the validation set, with a worst-group accuracy of 30.8\%. The most confusing class for baby pacifier is \emph{water bottle}. For all methods, we then fine-tune the CLIP RN50 model with the training data from these two classes: \emph{baby pacifier} and \emph{water bottle} to understand if such an isolated mitigation could positively align the model. In \cref{tab:imagenetresults} we see that in terms of both average accuracy and worst-group accuracy, the baseline ERM method performs just as well as our methods. However, since the test dataset in this case is rather small (only 50 images per class), it is useful to also look at the alignment of explanations. Figure~\ref{fig:imagenet-rn50-mitigated} illustrates this visually, highlighting that GradCAM maps are not focused on the baby face for our approach, which is the case for other methods. The same result is confirmed by the higher AIoU scores.

\textbf{Limitations.} While the method proposed here shows promising results for mitigating spurious correlations, learning pipelines often face a combination of problems that go beyond spurious features and involve other out-of-distribution shifts. We illustrate these concerns through a running example from ImageNet in Appendix~\ref{sec:limitations} and show that current decorrelation methods may not be sufficient when models deal with issues such as high concept variation, insufficient data, label noise, or visual commonalities between spurious and non-spurious features. 
\section{Conclusion and Future Work}
We proposed a language-based approach to mitigate spurious correlations of CLIP, as a contrastive learning vision and language model. Our focus on mitigations that can be initiated through language is motivated by the fact that spurious attribute annotations may not always be available. The contrastive loss function formulation guiding the spurious attribute decorrelation is applied at fine-tuning time and is effective even when the language and image encoders are excluded from the fine-tuning process. Besides the computational convenience, this is a promising finding speaking to the foundational nature of the larger representations. The work opens up several questions for future research, including the scalability of such methods when mitigating several spurious correlations at the same time. While this work focused on spurious correlations for classification tasks, studying the problem from a representational bias perspective and how spurious correlations may feed issues in representation fairness is an important relevant direction with several societal implications. Finally, we see opportunities in further leveraging model multi-modality and language-initiated mitigation actions to either generate teaching samples for mitigations or high-level instructions for the model to follow.

\textbf{Acknowledgment.}
This research was partially supported by Cisco Systems and the National Science Foundation CAREER
Award 2146492.

\bibliography{icml_CLIPSpurious}
\bibliographystyle{icml2023}

\newpage
\appendix
\onecolumn

\section{Ablation Study}
To better understand the impact and necessity of each loss term we conducted ablation studies for different combinations of components. \cref{tab:ablation_study} summarizes the findings. First, we see that choosing between the spurious image loss (row 2) or the spurious language loss (row 3) leads to similar results in worst-group accuracy but AIoU scores are higher for the spurious image loss, perhaps because decorrelation directly in the image space, but of course this requires attribute annotations. Second, the image contrastive loss is necessary for any of the spurious losses to be effective. For example, when we compare rows 2 and 3 with their corresponding versions in rows 4 and 5 that do not have the contrastive image loss, we see both worst-group acuraccy and AIoU decreasing. While this dependency on the contrastive image loss calls for more investigation, it also shows evidence that for improving vision classification results, the process of separating spurious attributes needs to happen hand in hand with separating classes from each other.
\label{sec:loss-ablation}

\begin{table*}
  \centering
    \caption{Investigating the impact and necessity of different loss terms with CLIP-RN50 and Waterbirds. The last four columns show the AIoU scores on the four groups (e.g. WB-L is waterbird with a land background.)
  }
  \begin{tabular}{@{}lcccccccc@{}}
    \toprule
     & \multicolumn{2}{c}{Accuracy}  & \multicolumn{2}{c}{AIoU} & \multicolumn{4}{c}{AIoU} \\

    $\mathcal{L}_{\text{CLIP}}$+ & Avg.  & Worst-group  & Avg. & Worst-group & LB-L & LB-W & WB-L & WB-W \\ \midrule
    (1) $\mathcal{L}_{lc}$+$\mathcal{L}_{vc}$+$\mathcal{L}_{vs}$+$\mathcal{L}_{ls}$ & 86.9\% & \textbf{78.2}\% & 0.550 & 0.281 & 0.244 & 0.281 & 0.852 & 0.823\\
    (2) $\mathcal{L}_{lc}$+$\mathcal{L}_{vc}$+$\mathcal{L}_{vs}$ & 83.2\% & \textbf{77.5}\% & \textbf{0.654} & \textbf{0.587} & 0.601 & 0.587 & 0.714 & 0.715\\
    (3) $\mathcal{L}_{lc}$+$\mathcal{L}_{vc}$+$\mathcal{L}_{ls}$ & 84.7\% & \textbf{77.5}\% & \textbf{0.628} & \textbf{0.499} & 0.460 & 0.499 & 0.788 & 0.764\\
    (4) $\mathcal{L}_{lc}$+$\mathcal{L}_{vs}$ & 83.8\% & 72.0\% & 0.525 & 0.310 & 0.329 & 0.310 & 0.762 & 0.697 \\
    (5)  $\mathcal{L}_{lc}$+$\mathcal{L}_{ls}$ & \textbf{88.0}\% & 75.1\% & 0.608 & 0.493 & 0.446 & 0.493 & 0.776 & 0.719 \\
    (6) $\mathcal{L}_{vc}$+$\mathcal{L}_{vs}$ & 86.8\% & 76.1\% & 0.609 & 0.461 & 0.441 & 0.461 & 0.768 & 0.763 \\
    \midrule
    Pre-trained CLIP & \textbf{90.8}\% & 44.9\% & 0.507 &  0.479 & 0.471 & 0.479 & 0.541 & 0.537 \\
    Fine-tuned CLIP & 81.3\% & 77.1\% & 0.510 &  0.128 & 0.098 & 0.128 & 0.929 & 0.885\\ 
    ERM & 93.5\% & 54.4\% & 0.514 & 0.139 & 0.187 & 0.139 & 0.920 & 0.919 \\
    Group DRO & 83.3\% & 73.7\% & 0.509 & 0.274 & 0.219 & 0.274 & 0.800 & 0.741\\
    \bottomrule
  \end{tabular}
  \label{tab:ablation_study}
\end{table*}

\begin{table}
  \centering
    \caption{Results of fine-tuning CLIP-RN50 with a subset of ImageNet classes, ``can opener" and ``letter opener". Both average and worst-group performance are evaluated with models early stopped at the \textit{highest worst-group accuracy} on the validation set.}
  \begin{tabular}{@{}lcccc@{}}
    \toprule
     Class x   & \multicolumn{2}{c}{Accuracy} & \multicolumn{2}{c}{AIoU}  \\
     Can Opener & Avg. & Worst & Avg. & Worst \\
    \midrule
    Pre-trained & 86.4\% & 75.0\% & 0.568 & \textbf{0.464} \\
    Fine-tuned  & 76.8\% & 68.0\% & 0.561 & 0.339 \\
    \midrule
    ERM & 80.1\% & 68.0\% & 0.436 & 0.295 \\
    Group DRO & 78.4\% & 76.0\% & 0.426 & 0.205 \\
    Ours($\mathcal{L}_{lc}$+$\mathcal{L}_{vc}$+$\mathcal{L}_{ls}$) & 73.9\% & 68.0\% & \textbf{0.585} & 0.344  \\
    Ours($\mathcal{L}_{lc}$+$\mathcal{L}_{vc}$+$\mathcal{L}_{vs}$) & 75.7\% & 72.0\% & \textbf{0.612} & \textbf{0.356}  \\
    \bottomrule
  \end{tabular}

  \label{tab:imagenet-can}
\end{table}

\section{Limitations}
\label{sec:limitations}
While our method has demonstrated promising results in mitigating spurious correlations in ImageNet, there are still limitations to consider. We have identified several circumstances where our method may not be effective, which we discuss in more detail below. 

\subsection{Spurious-aware Contrastive Language
Image Fine-tuning}
\begin{figure}[t]
\vskip 0.2in
\begin{center}
\centerline{\includegraphics[width=0.5\columnwidth]{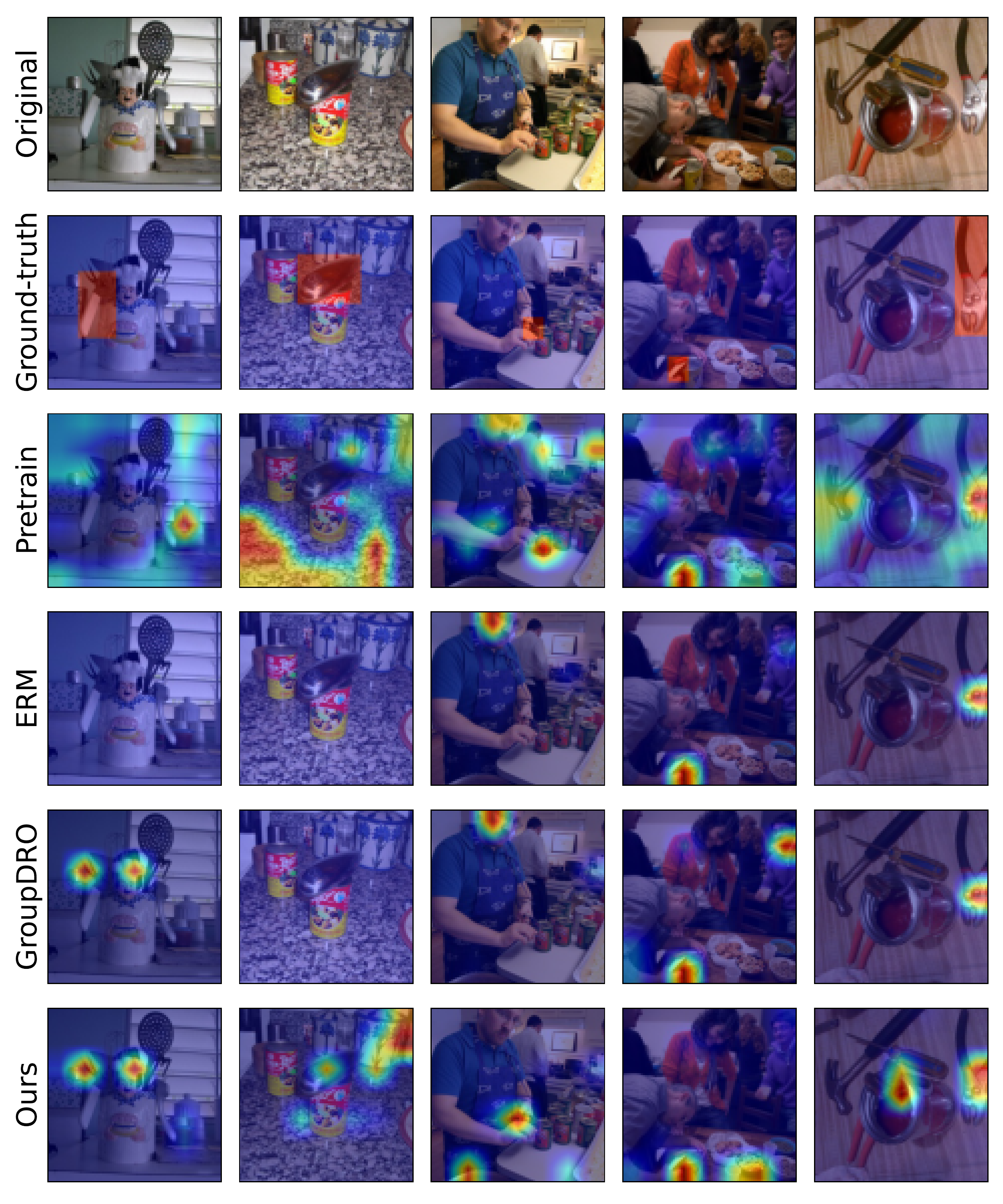}}
\caption{GradCAM explanations for different approaches based
on CLIP RN50 for the ImageNet dataset, ``can opener" class.}
\label{fig:imagenet-canopener}
\end{center}
\vskip -0.2in
\end{figure}

We will describe these limitations by using the example presented in Figure~\ref{fig:imagenet-canopener}. In this example, the target class we are studying is ``can opener" and the spurious feature being identified by our method is the ``can" itself. The spurious correlation leads to a 45.2\% discrepancy in accuracy and as shown in Figure~\ref{fig:imagenet-spurious}, the model often focuses on the can rather than the opener.

\begin{figure}[t]
\vskip 0.2in
\begin{center}
\centerline{\includegraphics[width=0.7\columnwidth]{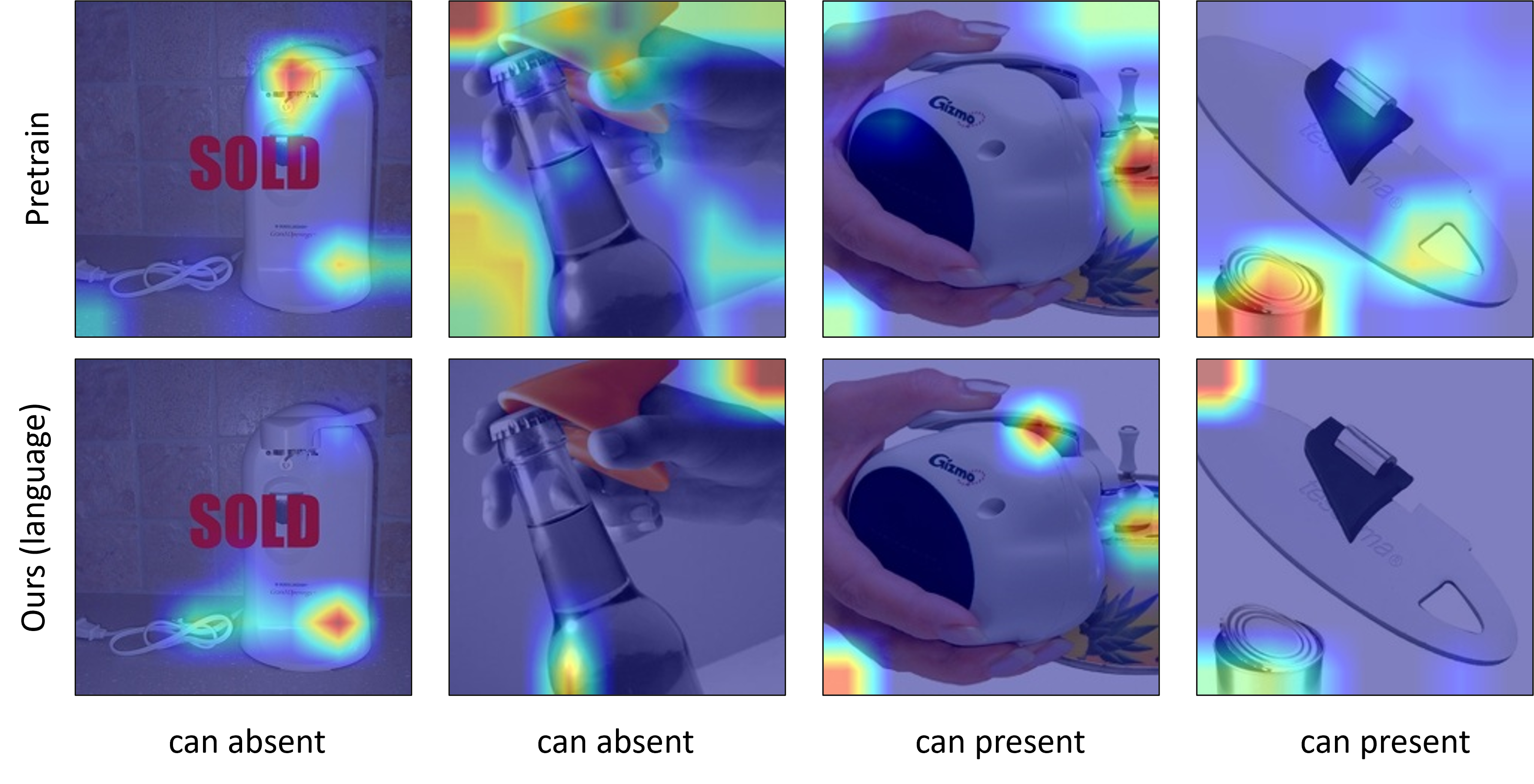}}
\caption{The ``can opener" class in ImageNet has several concept variations, posing additional challenges in the learning process. The presence of spurious correlations can in addition exacerbate these issues or even hide them when models are right for the wrong reasons (e.g., the last two examples in this figure). While our method reduces the focus on ``cans", it is still not able to completely alleviate the problem for examples with unusual concept variation.}
\label{fig:imagenet-canopener-concept-variation}
\end{center}
\vskip -0.2in
\end{figure}

\paragraph{High Concept Variation + Insufficient Data.} One limitation is that the effectiveness of our method can depend on the level of variation within each class. For example, some classes may have many different variations, such as different breeds of dogs or types of flowers, while other classes may have less variation, such as types of paperclips or staplers. 
It is essential to note that increased variation within a class can pose a significant challenge in the learning process, even without spurious correlations. The presence of spurious correlations however can exacerbate these issues or even hide them when models are right for the wrong reasons. For example, the model can still be correct on an image with an unusual class concept variation when the spurious correlation is present (see the last two examples in Figure~\ref{fig:imagenet-canopener-concept-variation}). Therefore, while correcting spurious correlations is vital, it alone may not address the inherent problem of object variation but the identification stage for spurious correlations itself may still be informative. Additionally, if there is insufficient data for a particular variation, our method may be unable to learn that variation effectively. For example, Figure~\ref{fig:imagenet-canopener-concept-variation}) shows that even though our approach tends to put less emphasis on the ``can" itself, it still cannot completely alleviate the problem or guarantee that the focus of the model will not shift to other parts of the image that are still spurious.

The same discussion applies to label noise, which can also be compounded by increased class variation or label subjectivity~\cite{pmlr-v119-shankar20c}.

\begin{figure}[t]
\vskip 0.2in
\begin{center}
\centerline{\includegraphics[width=0.6\columnwidth]{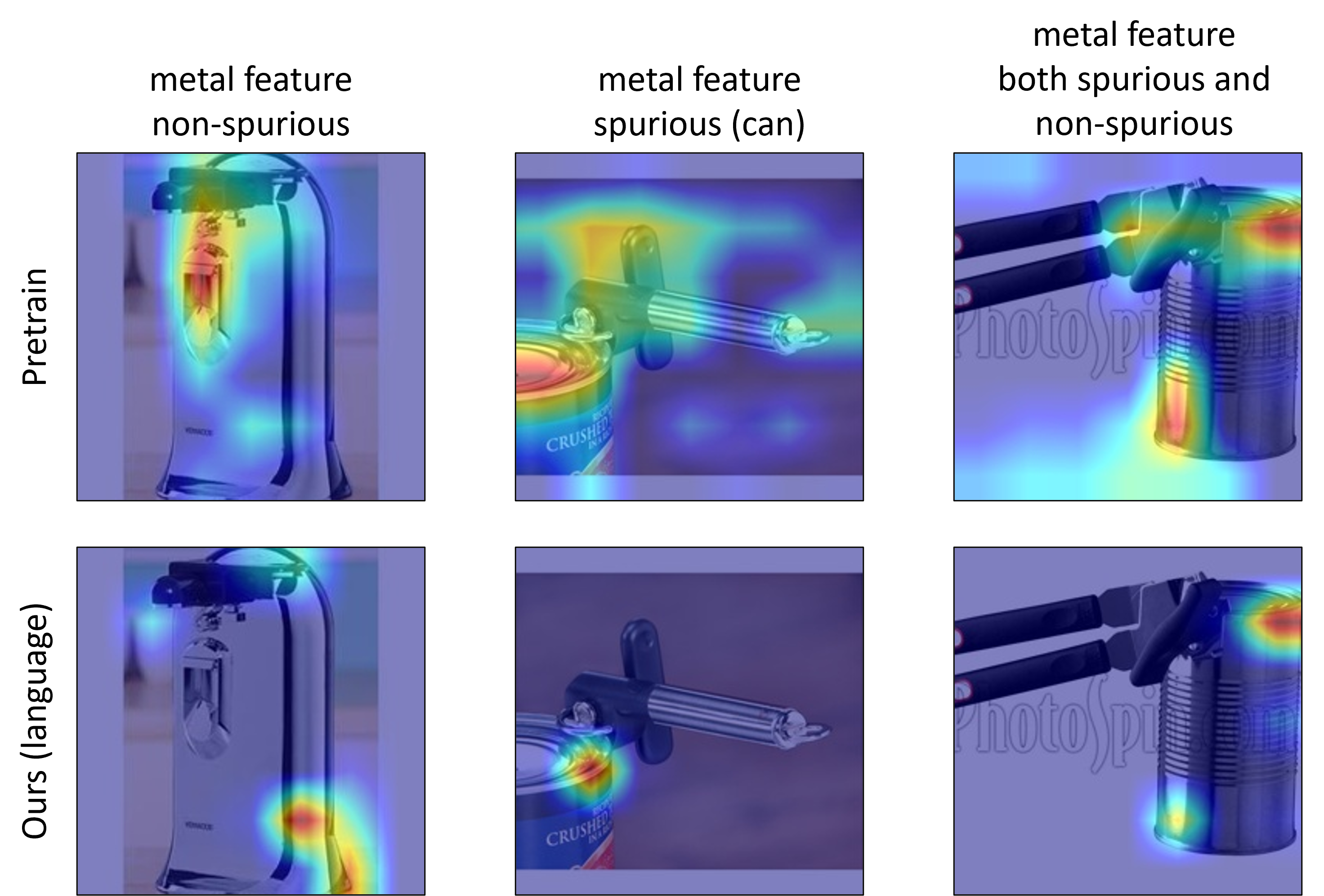}}
\caption{Metal features of can openers share visual commonalities with cans, which makes the problem of mitigating spurious correlations more difficult for such cases.}
\label{fig:imagenet-canopener-feature-definition-challenges}
\end{center}
\vskip -0.2in
\end{figure}

\paragraph{Feature Definition Challenges.} Our method relies on being able to identify and isolate spurious features from non-spurious ones. However, in some cases, the spurious feature may be difficult to define or may have parts that are similar to non-spurious features. For example, in the case of a can and a can opener, the metal parts of each object may look similar and contain similar features (Figure~\ref{fig:imagenet-canopener-feature-definition-challenges}). This can make it more challenging to separate the spurious correlations from the legitimate ones.

Due to these challenges in combination, in Table~\ref{tab:imagenet-can} we see that almost all methods, including ours, are not able to improve the accuracy for the ``can opener" class. It is important to note that these limitations are not unique to our method, but rather are common issues that arise with spurious correlation mitigation methods and their interactions with general learning objectives. Nevertheless, the approach proposed in this work can be beneficial in many cases, in particular when significant performance drops are mainly due to spurious correlations and when spurious features do not share a visual commonality with non-spurious ones. Further research is needed to address these limitations and improve the effectiveness of spurious correlation mitigation methods in combination with other forms of out-of-distribution shifts on concept variation.

\subsection{Spurious Correlation Detection}
One limitation of our spurious correlation detection method is that it relies on the object detector to identify relevant attributes. However, certain factors such as lighting and contrast may not be captured by the object detector, which can limit the effectiveness of our approach. To mitigate this limitation, our method can be used in conjunction with previous work that utilizes system/content metadata or discovered visual features for general failure analysis \cite{nushi2018towards, singlaCVPR2021, chung2019slice, jain2022distilling, eyuboglu2022domino}. For instance, one can enrich the test data with additional meta-data, such as contrast, blur, lighting, and camera angle, and apply our method as well as previous approaches to understand if the model's performance drops for some of these conditions. 

Note that our method leverages language attributes to describe spurious correlations but does not exclude the use of other metadata for the same purpose. While language attributes can capture a broad range of failure modes, incorporating all relevant information is crucial for comprehensive debugging.

\end{document}